\documentclass[journal]{IEEEtranTIE}
\usepackage{graphicx}
\usepackage{cite}
\usepackage{picinpar}
\usepackage{amsmath}
\usepackage{url}
\usepackage{flushend}
\usepackage[latin1]{inputenc}
\usepackage{colortbl}
\usepackage{soul}
\usepackage{multirow}
\usepackage{pifont}
\usepackage{color}
\usepackage{alltt}
\usepackage[hidelinks]{hyperref}
\usepackage{enumerate}
\usepackage{siunitx}
\usepackage{breakurl}
\usepackage{epstopdf}
\usepackage{pbox}

\def\BibTeX{{\rm B\kern-.05em{\sc i\kern-.025em b}\kern-.08em
		T\kern-.1667em\lower.7ex\hbox{E}\kern-.125emX}}

\begin{document}
\title{	\textit{In vivo} validation of Wireless Power Transfer System for Magnetically Controlled Robotic Capsule Endoscopy}

\author{
	\vskip 1em
	
	Alessandro Catania, \IEEEmembership{Senior Member, IEEE},
	Michele Bertozzi, Nikita J. Greenidge, Benjamin Calme, Gabriele Bandini, Christian Sbrana      \IEEEmembership{Student Member, IEEE}, Roberto Cecchi, Alice Buffi, \IEEEmembership{Senior Member, IEEE}, Massimo Macucci, \IEEEmembership{Member, IEEE}, Sebastiano Strangio \IEEEmembership{Senior Member, IEEE}, Pietro Valdastri, \IEEEmembership{Senior Member, IEEE}, and Giuseppe Iannaccone \IEEEmembership{Fellow, IEEE}

\thanks{\textbf{This work has been submitted to the IEEE for possible publication. Copyright may be transferred without notice, after which this version may no longer be accessible.} This work was supported by the EC Horizon 2020 Research and Innovation Programme under GA AUTOCAPSULE 952118 and partially supported by the italian MUR under the Forelab project of the ``Dipartimenti di Eccellenza'' programme.}
\thanks{A. Catania, R. Cecchi, M.Macucci and S. Strangio are with the Department of Information Engineering, University of Pisa, Italy (e-mail: alessandro.catania@unipi.it; roberto.cecchi@phd.unipi.it; massimo.macucci@unipi.it; sebastiano.strangio@unipi.it). }
\thanks{M. Bertozzi is with Clavis s.r.l., Pisa, Italy, Quantavis s.r.l., Pisa, Italy, and with the Department of Information Engineering, University of Pisa, Italy (e-mail: michele.bertozzi@quantavis.com)}
\thanks{C. Sbrana and G. Iannaccone are with Quantavis s.r.l., Pisa, Italy, and with the Department of Information Engineering, University of Pisa, Italy (e-mail: christian.sbrana@quantavis.com; giuseppe.iannaccone@unipi.it).}
\thanks{G. Bandini and A. Buffi are with the Department of Energy Systems Territory and Constructions Engineering, University of Pisa, Italy (e-mail: gabriele.bandini@ing.unipi.it; alice.buffi@unipi.it).}
\thanks{N.J. Greenidge, B. Calme and P. Valdastri are with the STORM Lab, Institute of Autonomous Systems and Sensing (IRASS), School of Electronic and Electrical Engineering, University of Leeds, U.K (e-mail: elnjg@leeds.ac.uk; b.p.calme@leeds.ac.uk; p.valdastri@leeds.ac.uk).}
}

\maketitle
	
\begin{abstract}
This paper presents the in vivo validation of an inductive wireless power transfer (WPT) system integrated for the first time into a magnetically controlled robotic capsule endoscopy platform. The proposed system enables continuous power delivery to the capsule without the need for onboard batteries, thus extending operational time and reducing size constraints. The WPT system operates through a resonant inductive coupling mechanism, based on a transmitting coil mounted on the end effector of a robotic arm that also houses an external permanent magnet and a localization coil for precise capsule manipulation. To ensure robust and stable power transmission in the presence of coil shift, misalignment or rotation, a 3D receiving coil and a closed-loop adaptive control system are implemented. The system has been extensively characterized in laboratory settings and validated through in vivo experiments using a porcine model, demonstrating reliable power transfer and effective robotic navigation in realistic gastrointestinal conditions: the average received power was 110~mW at a distance of 9~cm between the coils, with variable capsule rotation angles.
The results confirm the feasibility of the proposed WPT approach for autonomous, battery-free robotic capsule endoscopy, paving the way for enhanced diagnostic in gastrointestinal medicine.
\end{abstract}


\markboth{IEEE TRANSACTIONS ON INDUSTRIAL ELECTRONICS}%
{}

\definecolor{limegreen}{rgb}{0.2, 0.8, 0.2}
\definecolor{forestgreen}{rgb}{0.13, 0.55, 0.13}
\definecolor{greenhtml}{rgb}{0.0, 0.5, 0.0}

\section{Introduction}
\label{sec:introduction}

Colonoscopy procedures are vital for the early detection and diagnosis of colorectal diseases, including colorectal cancer, one of the primary causes of cancer-related  fatalities globally \cite{Bray2024, Siegel2024}. Traditional colonoscopies, while effective, are often uncomfortable, painful and require extensive preparation, leading to a significant number of patients avoiding the procedure. Capsule endoscopes offer a patient-friendly alternative, as they explore the colon with minimal discomfort and pain, removing the need for sedation \cite{Oka2022}. 
A key innovation enabling the precise control and navigation of capsule endoscopes is the Magnetic Control System (MCS), introduced in \cite{Carpi2006}. The MCS operates through the interaction between a large External Permanent Magnet (EPM) and/or inductive coils, typically housed in a robotic arm or controller, and one or more miniaturized Internal Permanent Magnets (IPMs), embedded within the capsule. %The external magnetic field allows physicians to move the capsule to the area of interest and to adjust its position and orientation in real time and with precision\cite{Kim2022, Ye2022}. 
This technology not only enhances the accuracy and efficiency of the procedure but also contributes to improve patient comfort and accessibility, addressing many of the barriers associated with traditional colonoscopy methods. The integration of Wireless Power Transfer (WPT) with MCS into robotic capsule endoscopes has the advantage to allow both continuous power supply and precise navigation. The use of WPT eliminates the need for onboard batteries, which are typically limited by size, capacity, and safety constraints \cite{Cao2024}, hence extending the capsule operating time and allowing for longer and more extensive inspections. 
%Moreover, the use of a Wireless Data Transfer (WDT) system, together with WPT and MCS, paves the way for the vision of a fully-untethered capsule enabling the explorations of larger portions of the GI tract.

%An additional step forward would be the in-capsule integration of various sensing systems: imaging (both white light for traditional screening and navigation support, and micro-ultrasound for transmural or cross-sectional imaging of the colon walls \cite{Lay2019}), impedance spectroscopy, temperature and pH monitoring. This ambitious outcome cannot be achieved without the aggressive miniaturization and integration of all subsystems and the enhancement of their power efficiency.

The use of inductive WPT in Implantable Medical Devices (IMDs) has been widely explored, for its potential to radically expand the number of addressable clinical use cases and applications\cite{Haerinia2020,Khan2020,Zhou2020}. %For instance, inductive WPT systems for powering cardiac implants \cite{Zhang2022} and neurostimulators \cite{Biswas2023}, have been proposed, showcasing the benefits of eliminating battery limitations. 
The main constraint is that the transmitted power must be compliant to the Specific Absorption Rate (SAR) guidelines for human tissues \cite{ICNIRP2020}, which pose a limit to the power dissipated in the patient's body due to Joule effect. There are specific additional challenges in the use of WPT for capsule endoscopy: (i) the distance between Transmitting (Tx) and Receiving (Rx) coils is typically much larger than the size of the Rx coil itself, thus resulting in a very low magnetic coupling coefficient; (ii) the Rx coil position is not fixed as in other IMDs and the relative distance between Tx and Rx coils varies during the medical procedure;  (iii) coil misalignment and rotation can further negatively affect power transfer efficiency; (iv) the power budget required by the capsule, especially when employing multi-modal sensing systems, can be relatively high ($ > 100~\unit{\milli\watt}$).

%Previous studies have primarily focused on improving imaging capabilities and navigation of capsule endoscopes using magnetic control systems. While these advancements have significantly enhanced the diagnostic capabilities of capsule endoscopes, the issue of power supply has remained a persistent challenge.
In this paper we present - for the first time - a wireless-powered, magnetically controlled capsule endoscopy system, where the Tx coil for WPT is mounted on the end effector of the robotic arm that houses also the EPM responsible for manipulating and controlling the capsule. This configuration ensures that the capsule is constantly tracked and powered, regardless of its position within the colon. To further improve the robustness of the inductive link, a 3D coil system is integrated in the capsule to minimize the impact of misalignment and rotation, and to ensure reliable power delivery. The WPT receiver in the capsule has been miniaturized with custom designed integrated circuits, that include Full-Wave Rectifiers (FWRs) and circuits for Load-Shift-Keying (LSK) modulation. LSK modulation provides information about the received power back to the Tx side, allowing for an adaptive closed-loop control of the transmitted power. 
%This regulation is crucial for maintaining power transfer efficiency while adhering to the SAR limits for patient safety. 
Extensive characterization of the WPT receiver and the adaptive power control system has been conducted, followed by \textit{in vivo} animal trials. These tests have demonstrated the robustness of the proposed WPT system, capable of maintaining efficient power transfer in compliance with SAR guidelines. %This work represents a significant advancement in the functionality and reliability of robotic capsule endoscopy, and it has been developed within the AUTOCAPSULE Project \cite{Autocapsule}, i.e., Autonomous multimodal implantable endoscopic capsule for the gastrointestinal tract.

%The paper is structured as follows. Section \ref{sec:sys_overview} provides an overview of the robotic capsule endoscopy system, discussing the major challenges and the state of art, with a focus on the magnetic control system and the inductive WPT mechanism. Section \ref{sec:materials_methods} focuses on the design of the external transmitter and of the implantable WPT receiver. Section \ref{sec:results} presents the testing of the WPT system for a magnetically controlled robotic capsule endoscope, including the results obtained from \textit{in vivo} experiments. Finally, Section \ref{sec:conclusion} concludes the work and offers perspectives for future developments.

\section{Wirelessly Powered, Magnetically Controlled Capsule Endoscopy}
\label{sec:sys_overview}

%The design of endoscopic capsules requires the integration of several complex systems, including locomotion, advanced imaging and sensing, power management, data transmission, and onboard processing units \cite{Cao2024}. 
The focus of this work is on the WPT system and its integration with the MCS. These two technologies are crucial in enhancing the functionality and usability of fully untethered capsules, enabling for precise navigation through the gastrointestinal tract and ensuring a continuous power supply without the need for bulky and potentially chemical hazardous batteries. %Here, we will present a concise survey of the most popular approaches for these two systems.

\subsection{Magnetic Control System}

When selecting an actuation method for minimally invasive and untethered applications, power consumption and miniaturization are critical factors. Magnetic actuation is ideal as it enables safe, wireless and contact-free control while leaving onboard power available for essential components such as imaging, localization and data transmission.

Magnetic fields for capsule manipulation are typically generated using permanent magnets or electromagnetic coils. Permanent magnet systems \cite{xuAdaptiveSimultaneousMagnetic2022}, often employing a robotic arm-mounted magnet, are compact, energy-efficient, and capable of producing strong magnetic fields over a large workspace. However, they offer limited dynamic control over field strength and direction \cite{huoDesignControlClinical}. 
Electromagnetic coil systems \cite{sonMagneticallyActuatedSoft2020a, hoangIndependentElectromagneticField2021}, by contrast, use an array of coils to generate and modulate magnetic fields in real time. While they allow for precise control, their high power consumption and cooling requirements \cite{huoDesignControlClinical} make them less practical for prolonged clinical use. 

An additional consideration for capsule endoscopy is real-time localization to track position and orientation of the capsule within the gastrointestinal tract to enable closed-loop control. In this work, we build on the Magnetic Flexible Endoscope system \cite{obsteinMagneticFlexibleEndoscope2024}, which combines a robotic arm-mounted permanent magnet for actuation with an electromagnetic coil for localization, enabling precise real-time control while maintaining clinical usability. 

\subsection{Inductive Wireless Power Transfer}

%The non-radiative inductively coupled WPT is ideally suited for IMDs. It provides safe, efficient, and reliable power transfer without the need for physical connectors \cite{Khan2020}, thus eliminating the risk of infections and enhancing patient comfort. Inductive WPT operates safely using non-ionizing radiations and adhering to strict SAR regulations to minimize the risks of tissue heating. 

This technology supports miniaturized receiver designs that are suitable for compact, biocompatible implants, and enables continuous power delivery or recharging for long-term device usage, in order to reduce the need for surgical battery replacements. However, the power efficiency of inductive links is intrinsically sensitive to the distance, misalignment, and rotational orientation between Tx and Rx coils. This aspect is particularly pronounced in IMDs, owing to the need for miniaturized dimensions, and further aggravated in capsule endoscopes by the inherent anatomical constraints and movements during the medical procedure. The approaches to address these challenges mainly concern  coil design optimization, and adaptive control systems.

\textbf{Coil design optimization} involves the choice of materials, geometries and deployment in space, for both the Tx and the Rx coils. 
%Besides the standard approach of using litz wires to reduce the resistive losses due to skin effect at high frequency, the use of ferrite cores \cite{Wu2023} or of multi-layer self-resonant structures \cite{Stein2019} represents an effective solution to enhance the magnetic coupling and improve the coil quality factors, respectively. However, these approaches are mainly feasible on the Tx side, whereas the constraints of the implant require small and lightweight Rx coils, as the flexible, contact lens-like coil for ocular implants presented in \cite{Kim2015}. 
In particular, the evolution of coil geometries -- e.g. modifications of the standard Helmholtz coil \cite{Carta2011}, obtained by merging it with birdcage coil \cite{Basar2017, Campi2021}, as well as the development of saddle-shaped coil pair \cite{Zhuang2023} and butterfly-shaped coils \cite{Ha-Van2019} -- aims at enhancing the uniformity of the magnetic field generated by the Tx coil and the transmission stability. These advancements often come at the cost of increased design complexity and challenges during the integration with the MCS. Another popular approach consists in the use of a 3D-coil system, where each coil is aligned along one of the three orthogonal axes (x, y, z), to either generate \cite{Sergkei2018} or receive \cite{Lenaerts2007} the magnetic field, ensuring robust magnetic coupling regardless of the relative orientation between Tx and Rx.

\textbf{Adaptive control systems} complement these design improvements by dynamically adjusting the transmitter power to maintain optimal power transfer. It is worth reminding that the SAR is assessed as the rate at which electromagnetic energy is absorbed per unit mass (either for the whole body or a specific tissue mass), averaged over a 6-minute time window. This averaging accounts for the body's natural ability to dissipate heat over time, ensuring that short-term peaks in energy absorption do not result in excessive tissue heating. 
%This approach enables higher transmitted power during short periods of increased distance between the coils or during peaks of intense capsule activity. 
Using an adaptive control system, the transmitted power can be temporarily increased to meet these demands and subsequently reduced to ensure compliance with the SAR limits over the averaging period.
An important aspect of these control systems concerns the communication of the received power level back to the Tx side. An effective way to implement this communication exploits the same inductive link established for power transmission, using the backscatter principle \cite{Koelle1975}.
% similarly to RFID technology \cite{DeVita2005}. 
%Reliable, high data-rate transmission with LSK modulation has been demonstrated by connecting a switch in parallel with the load \cite{Lin2016}, at the cost of severe power penalties. More efficient strategies -- such as periodic switching between resonant and non-resonant conditions in current-mode rectifiers \cite{Yao2023} or the use of reconfigurable rectifiers \cite{Li2015} -- have been proposed but never demonstrated at transmission distances exceeding 1.5\si{cm}. 
Recently, we proposed a solution consisting of an LSK modulation system that maximizes both the power efficiency and the modulation depth by means of an adaptive matching network on the Rx side \cite{Bertozzi2023}.

\subsection{Proposed System}

\begin{figure}[!t]
\centerline{\includegraphics[width=1.0\columnwidth]{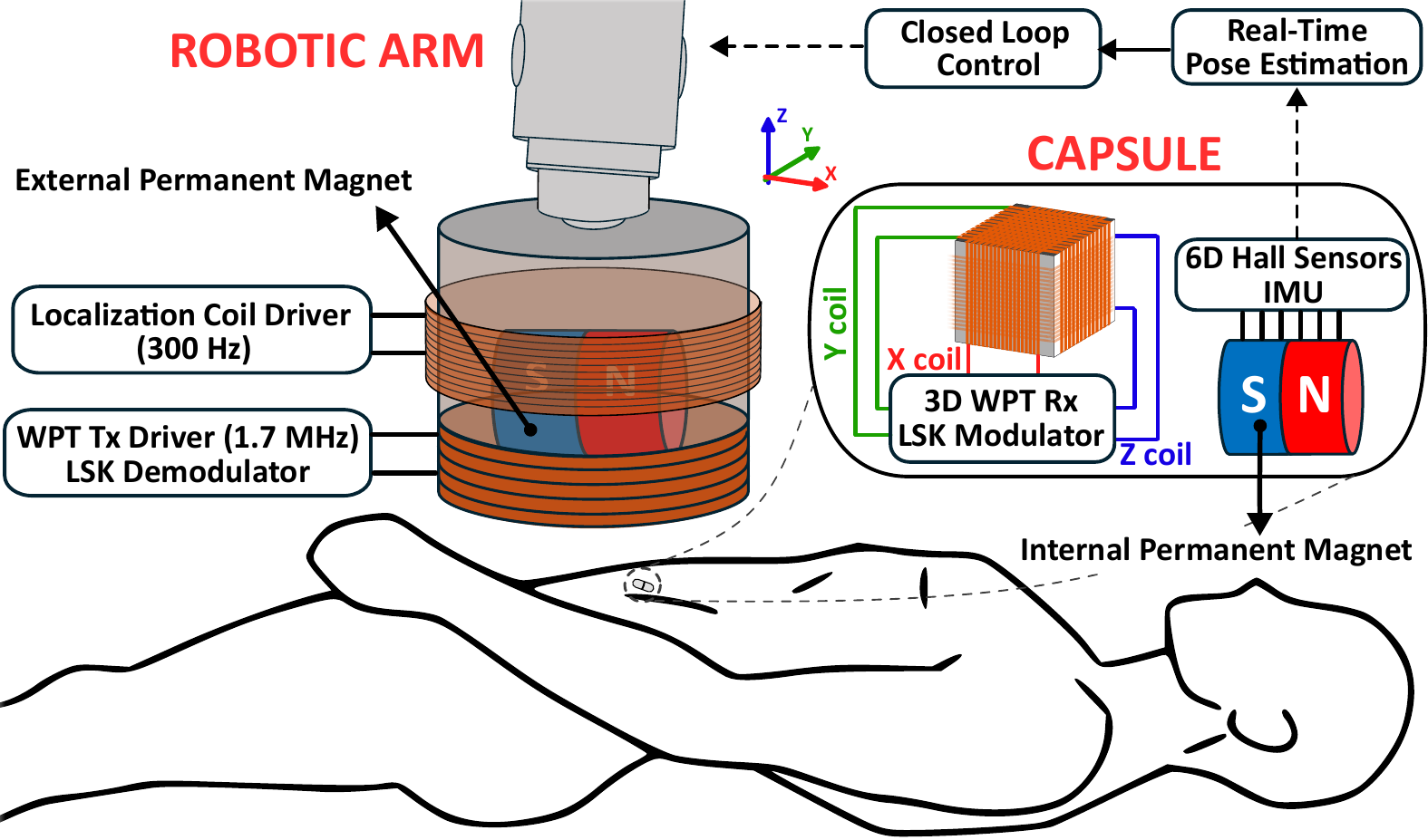}}
\caption{Proposed capsule endoscopy with closed loop MCS and 3D inductive WPT system.}
\label{fig2}
\end{figure}

The proposed capsule endoscopy system, depicted in Fig.~\ref{fig2}, includes for the first time both the MCS for closed-loop robotic magnetic manipulation and the WPT for continuous power delivery. The MCS comprises a hybrid magnetic field system able to combine the static field generated by an EPM and the time-varying field generated by a localization coil. Both the EPM and the coil are mounted on the end effector of a robotic arm, orthogonal to each other \cite{Taddese2018}. %The EPM is used to create magnetic force ($\textbf{F}$) and torque ($\boldsymbol{\tau}$) on the magnetic endoscope:
%\begin{equation}
%    \boldsymbol{\tau} = \mathbf{m} \times \mathbf{B}.
%    \label{torque}
%\end{equation}

%\begin{equation}
%    \textbf{F} = \nabla\mathbf{B}^T  \textbf{m}.
%    \label{force}
%\end{equation}
%where $\mathbf{m}$ is the magnetic moment of the IPM, $\mathbf{B}$ is the magnetic field generated by the EPM and $\nabla\mathbf{B}$ is its spatial gradient. 
In addition to using the forces and torques created by the EPM field to wirelessly steer the IPM within the capsule, this dual-source methodology allows a magnetic pose estimation method that addresses singularity problems in specific areas of the workspace, which previously led to a loss of estimation capability \cite{DiNatali2014}. The magnetic pose estimation requires also the information coming from 6D Hall-effect sensors (a pair of 3-axis Hall-effect sensors positioned at a known distance from each other) and an Inertial Measurement Unit (IMU), both integrated in the capsule alongside the IPM. The integration of this hybrid magnetic field system along with a particle filter-based state estimation framework eliminates the need for accurate initialization of the capsule yaw angle. The localization coil is driven by a square wave at 300~\unit{\hertz}, a frequency low enough to ensure adequate sampling for the localization algorithm and no absorption of the magnetic field by the human body. Full details on the MCS can be found in previous work \cite{Martin2022}, while the WPT system is deeply described in the next Section.

\section{Design of the Inductive WPT System}
\label{sec:materials_methods}

\begin{figure*}[!ht]
\centerline{\includegraphics[width=1\textwidth]{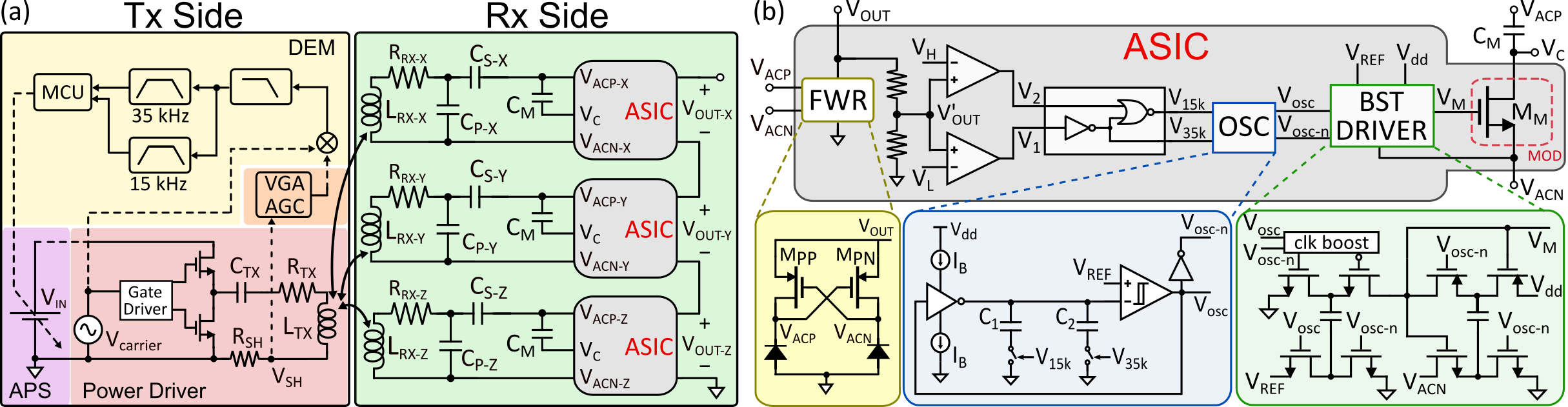}}
\caption{(a) Block diagram of the Wireless Power Transfer System: Tx side with Power Driver and LSK Demodulator, Rx side with 3D coil and ASICs dedicated to AC-DC conversion and LSK modulation. (b) Block diagram of the custom ASIC, with detailed schematic view of the FWR, the programmable oscillator (OSC) and the bootstrap (BST) driver.}
\label{fig3}
\end{figure*}

The complete block diagram of the proposed WPT system is depicted in Fig.~\ref{fig3}(a). The inductive link comprises the transmitter coil $L_{\rm TX}$, associated with its Equivalent Series Resistance (ESR) $R_{\rm TX}$, and the 3D receiver coil, consisting of $L_{\rm RX-X}$, $L_{\rm RX-Y}$ and $L_{\rm RX-Z}$, each characterized by its respective ESR, $R_{\rm RX-X}$, $R_{\rm RX-Y}$ and $R_{\rm RX-Z}$. Both the transmitter and the receiver work at resonance condition by means of properly sized capacitive matching networks in order to maximize the power efficiency \cite{Khan2020}. The Tx side consists of several separate subsystems that implement the power driving of the coil, the LSK demodulation and the adaptive control of the transmitted power. On the Rx side, the AC-DC converter and the LSK modulator are in three identical application-specific integrated circuits (ASICs)\cite{Rizzo2022}, one for each axis of the 3D coil. It is worth mentioning that, while the WPT receiver needs to be miniaturized due to the size constraints of the capsule, the Tx side does not face significant limitations in terms of volume occupancy.

%\begin{figure}[!t]
%\centerline{\includegraphics[width=1.0\columnwidth]{Figures/Figure3.png}}
%\caption{Block diagram of the Wireless Power Transfer System: Tx side with Power Driver and LSK Demodulator, Rx side with 3D coil and ASICs dedicated to AC-DC conversion and LSK modulation.}
%\label{fig3}
%\end{figure}

\subsection{Receiver Side}

Each coil $L_{\rm RX-i}$ (i=X,Y,Z) is connected to a parallel-series resonant matching network (made by $C_{\rm P-i}$ and $C_{\rm S-i}$) and then connected to a custom ASIC, which includes a full-wave rectifier and an LSK modulator. The load impedance modulation is implemented by means of an external capacitor $C_M$. The three AC-DC converters can be connected in series to obtain the final output voltage as depicted in Fig.~\ref{fig3}(a). 

Fig.~\ref{fig3}(b) illustrates the block diagram of the ASIC. The conversion of the input ac signal $(V_{\rm{ACP}}-V_{\rm{ACN}})$ into the dc output voltage ($V_{\rm{OUT}}$) is realized by a passive FWR with the cross-coupled connection between the MOSFETs $M_{\rm{PP}}$ and $M_{\rm{PN}}$ in order to implement low on-resistance switches, while diodes are employed in the lower part. 
%Replacing the diodes with active diodes brings to active rectifier topologies with higher power conversion efficiencies, but at the cost of increased design complexity due to the need for additional calibrations \cite{Namgoong2021}. In this work, we opted for a simpler yet less efficient FWR architecture, prioritizing the integration of the entire WPT system and reserving opportunities for further improvements in future works. 
The previously outlined adaptive control system requires the detection of the receiver output level and the modulation of the load impedance. These two operations are implemented in the ASIC. A partition of $V_{\rm{OUT}}$, called $V'_{\rm{OUT}}$, is compared with the thresholds $V_{\rm{H}}$ and $V_{\rm{L}}$ by means of two comparators with hysteresis, depending on the results of these comparisons, the frequency $f_{\rm{M}}$ generated by a programmable relaxation oscillator (OSC) is used to implement the LSK modulation. More in detail, the generated clock signal $V_{\rm{OSC}}$ is used to drive the nMOS switch $M_{\rm{M}}$, which periodically connects/disconnects the external capacitor $C_{\rm{M}}$, placed in parallel to the input of the FWR. The correct turn-on and turn-off operations are guaranteed by means of a bootstrap (BST) driver, that generates the gate signal $V_{\rm{M}}$ \cite{Rizzo2022}.
Depending on the level of $V'_{\rm OUT}$, three possible cases can be identified
\begin{itemize}
    \item for $V'_{\rm OUT} < V_L$, no modulation ($f_M = 0~\unit{\hertz})$;
    \item for $V_L < V'_{\rm OUT} < V_H$, modulation at $f_M = 15~\unit{\kilo\hertz}$;
    \item for $V'_{\rm OUT} > V_H$, modulation at $f_M = 35~\unit{\kilo\hertz}$.
\end{itemize}
The periodic connection of $C_{\rm{M}}$ alters the reflected impedance on the Tx side, resulting in a fluctuation of the amplitude of the current flowing in the coil, which can be detected in order to close the adaptive control loop, as further clarified in the next Section.

\subsection{Transmitter Side}

The Tx side includes several subsystems, as depicted on the left side of Fig.~\ref{fig3}(a). The Power Driver (working at $f_0 = 1.7~\unit{\mega\hertz}$) consists of a half-bridge inverter loaded by the series connection of the coil $L_{\rm TX}$ and the resonant capacitor $C_{\rm{TX}}$, characterized by a high quality factor. The voltage across the shunt resistor $R_{\rm{SH}}$, namely $V_{\rm{SH}}$, placed in series with the Tx coil, is then proportional to the current flowing in $L_{\rm{TX}}$. Considering that the input voltage of the Power Driver $V_{\rm{IN}}$, provided by an Adjustable Power Supply (APS), can cover a wide range, also $V_{\rm{SH}}$ will witness the same relative variations. For this reason, a Variable Gain Amplifier (VGA) with Automatic Gain Control (AGC) regulates the amplitude of $V_{\rm{SH}}$ to not exceed the limits of the LSK Demodulator (DEM). 

The output of the VGA is mixed with the carrier signal $V_{\rm{carrier}}$ and then low-pass filtered to extract the LSK modulated signal. Two selective band-pass filters, centered at 15~\rm{kHz} and at 35~\unit{\kilo\hertz}, recognize the presence of the respective modulation performed by the LSK modulator on the Rx side. This information is used by the Micro-Controller Unit (MCU) to control the APS output voltage and consequently the transmitted power, thereby maintaining the closed-loop control of the WPT system. For instance, if neither filter detects the modulation, it indicates that the rectified voltage is below the lowest threshold, i.e., the received power level is insufficient. Consequently, the input voltage $V_{\rm{IN}}$ is increased, leading to a higher current through the Tx coil and a corresponding enhancement of the generated magnetic field. This adjustment is performed iteratively by a control algorithm implemented in the MCU, until the power level at the Rx is adequate. Once the power level reaches the desired range, the LSK modulator begins operating at 15~\unit{\kilo\hertz}, and the LSK demodulator detects this signal, stopping further increasing of $V_{\rm{IN}}$. Similarly, if the received power level is too high, i.e., when the modulated frequency is 35~\unit{\kilo\hertz}, a corresponding adjustment is made to reduce the input voltage, ensuring that the power level remains within an acceptable range.

\section{Experiments and In-vivo trials}
\label{sec:results}

\subsection{Demonstrator of the WPT System}

Given the unique application-specific requirements, both the Tx and Rx coils and the receiver have been custom designed, while off-the-shelf components were used for the transmitter wherever feasible, since the constraints regarding area and power consumption are less stringent.
Fig.~\ref{fig5}(a) shows all the printed circuit boards (PCBs) employed on the Tx, exploiting the following components and systems:

\begin{itemize}
    \item the EPC90135 development board, a 100--\unit{\volt}, 25--\unit{\ampere} Half-Bridge Driver with GaN field-effect transistors employed as part of the Power Driver;
    \item the AD8338-EVALZ board, which implements the AGC on AD8338 Low Power VGA; 
    \item the Arduino Uno chosen as the microcontroller board (based on the ATmega328P MCU) to interface the DEM board with the APS.
\end{itemize}

The Power Driver board includes also: (i) a tunable oscillator that generates the carrier signal $V_{\rm{carrier}}$ for the half-bridge gate driver, (ii) a shunt resistor, obtained connecting six 0.4-\unit{\ohm} resistors in parallel, and (iii) the series-capacitance matching network on the Tx side ($C_{\rm{TX}}$ in Fig.~\ref{fig3}(a)), implemented using a series-parallel arrangement of high-voltage mica capacitors capable of withstanding 2500~\unit{\volt}. This configuration is designed to distribute the resonance voltage -- potentially high due to the elevated $Q$ factor of the Tx coil -- across multiple capacitors, thereby enhancing reliability and reducing the risk of overvoltage stress. The DEM board uses the AD835 mixer and banks of 4\textsuperscript{th}-order Chebyshev filters in multiple-feedback configuration. The APS is based on the LM76005, a 3.5~\unit{\volt} to 60~\unit{\volt}, 5~\unit{\ampere} synchronous step-down converter, with a programmable feedback network to update the voltage conversion ratio, as required by the adaptive control system.

Fig.~\ref{fig5}(b) shows the measurement setup with the Tx and the 3D Rx coil. The Tx coil has a diameter of 20~\unit{\centi\meter} and 6 turns of litz wires (2000 wires, 50~\unit{\micro\meter} diameter each one), resulting in an inductance of $L_{\rm TX}=16.7~\unit{\micro\henry}$, and an ESR $R_{\rm TX}=0.73~\unit{\ohm}$. This yields to a quality factor $Q_{\rm TX}$ of 244 at $f_0 = 1.7~\unit{\mega\hertz}$. The 3D coil has a cube shape of 1~\unit{\cm}-side with three hand-wound windings, one for each axis, made with enameled copper wire with a diameter of 250~\unit{\micro\meter}. $L_{\rm RX-i}$ and $R_{\rm RX-i}$ are approximately 6.5~\unit{\micro\henry} and 2.9~\unit{\ohm}, respectively, with a variation of less than 3\% across the three axes, leading to a quality factor $Q_{\rm RX}$ of 23.9. 
The parallel-series resonant matching networks on the Rx side are designed to resonate at 1.7~\unit{\mega\hertz} and maximize the power efficiency for the heaviest load (120~\unit{\ohm}), then $C_{\rm P-i} = 1.12~\unit{\nano\farad}$ and $C_{\rm S-i} = 120~\unit{\pico\farad}$. The surface-mount-device capacitors, together with $C_{\rm{M}}$ of $280~\unit{\pico\farad}$, were bonded on the three PCBs in Fig.~\ref{fig5}(b) (one for each axis), specifically designed to fit inside the capsule prototype and minimize the area consumption (smaller than 
$12 \times 15$ \unit{\milli\meter}\textsuperscript{2}). Each PCB also hosts the custom ASIC, mounted in a QFN48 package (size 7 x 7 \unit{\milli\meter}\textsuperscript{2}). The 120-\unit{\ohm} load represents the heaviest load condition when considering a target power budget of 100~\unit{\milli\watt} obtained with a rectified output voltage of 3.3~\unit{\volt}. The optical micrograph of the ASIC is shown in Fig.~\ref{fig5}(c). It was fabricated with a 180~\unit{\nano\meter} BCD-on-SOI CMOS process on a $1.5 \times 1.5$~\unit{\milli\meter}\textsuperscript{2} area.

\begin{figure}[!b]
\centerline{\includegraphics[width=0.9\columnwidth]{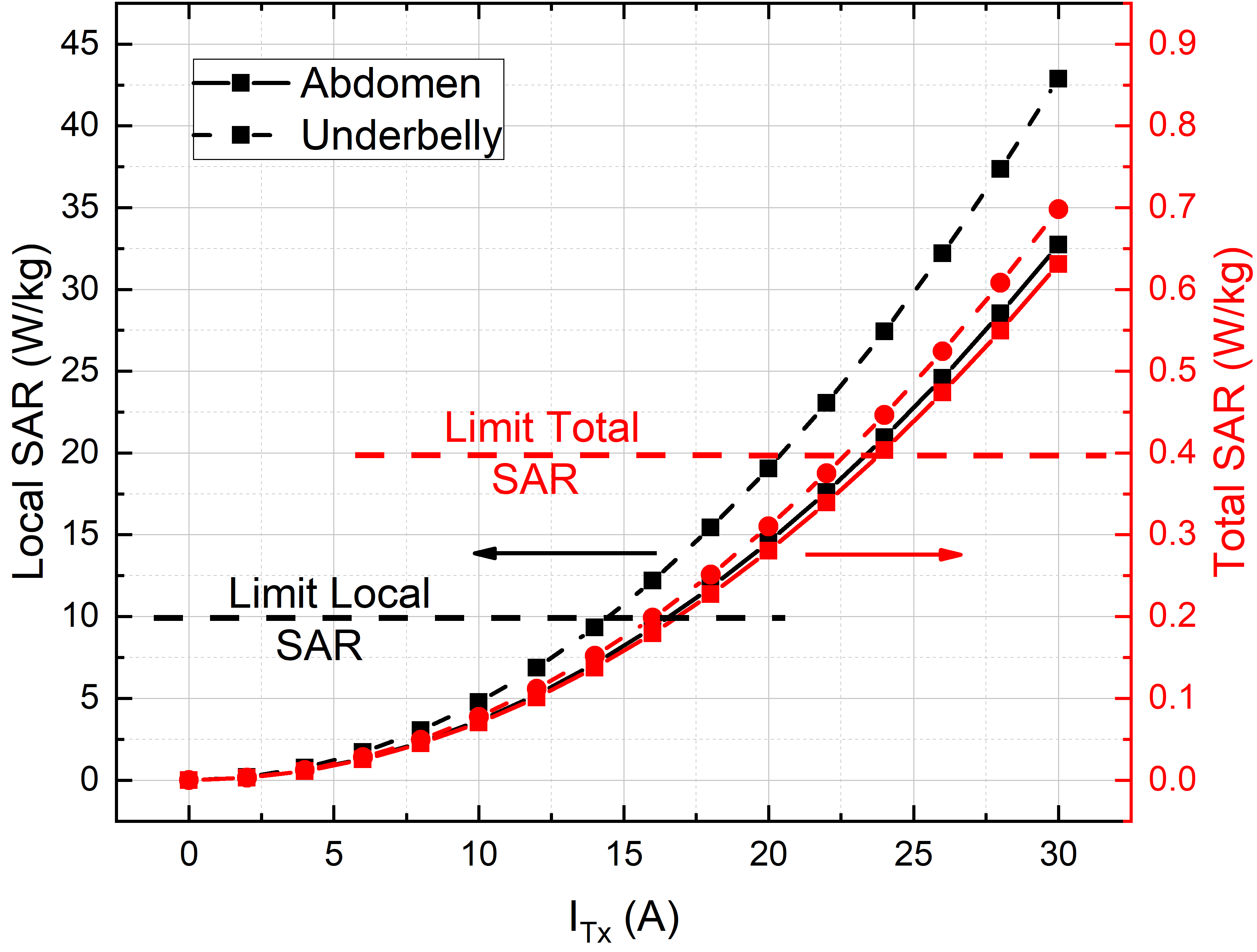}}
\caption{Total and local SAR estimated from 3D electromagnetic simulations (performed with COMSOL Multiphysics software) on a human body model. The TX coil was placed above abdomen and underbelly (at 3~\unit{\centi\meter} from the skin) and total and local SAR were measured for different amplitudes of the sinusoidal current on the Tx coil at 1.7 MHz.}
\label{figSAR}
\end{figure}

\begin{figure*}[!ht]
\centerline{\includegraphics[width=1\textwidth]{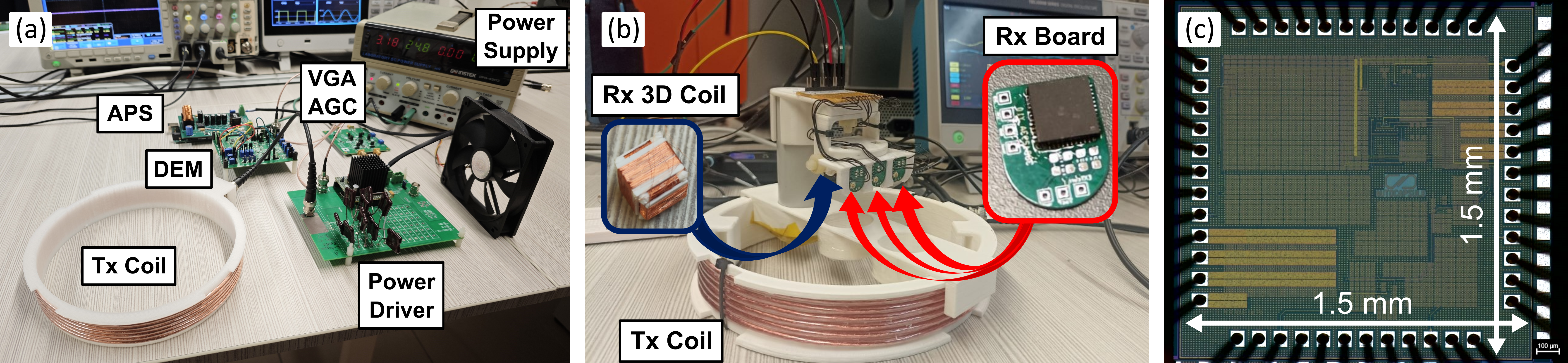}}
\caption{(a) Experimental setup of the proposed WPT transmitter. (b) Tx coil and 3D Rx coil connected to the 3 miniaturized PCBs, hosting the capacitive matching network, the modulation capacitance, and the ASIC integrating the rectifier and LSK modulator. (c) Optical micrograph of the ASIC.}
\label{fig5}
\end{figure*}

\subsection{Characterization of the WPT System}
3D electromagnetic simulations with the finite element simulation software COMSOL Multiphysics have been performed on a human-body model, with the Tx coil placed above the abdomen and underbelly to assess the dissipated power in the human body at 1.7~\unit{\mega\hertz} (Fig.~\ref{figSAR}).  At a distance of 3~\unit{\cm} from the skin, the most restrictive condition was dictated by the local SAR \cite{ICNIRP2020}, i.e. the average of body losses on the most exposed 10-g cubic mass over 6-minutes time interval. This constraint resulted in a maximum constant current amplitude flowing in the Tx coil of 14.5~\unit{\ampere}. This value has been taken as the upper bound limit during \textit{in vivo} tests. Nonetheless, higher values could be tolerated for shorter durations by employing the previously proposed adaptive control mechanism.

\begin{figure}[!b]
\centerline{\includegraphics[width=1.0\columnwidth]{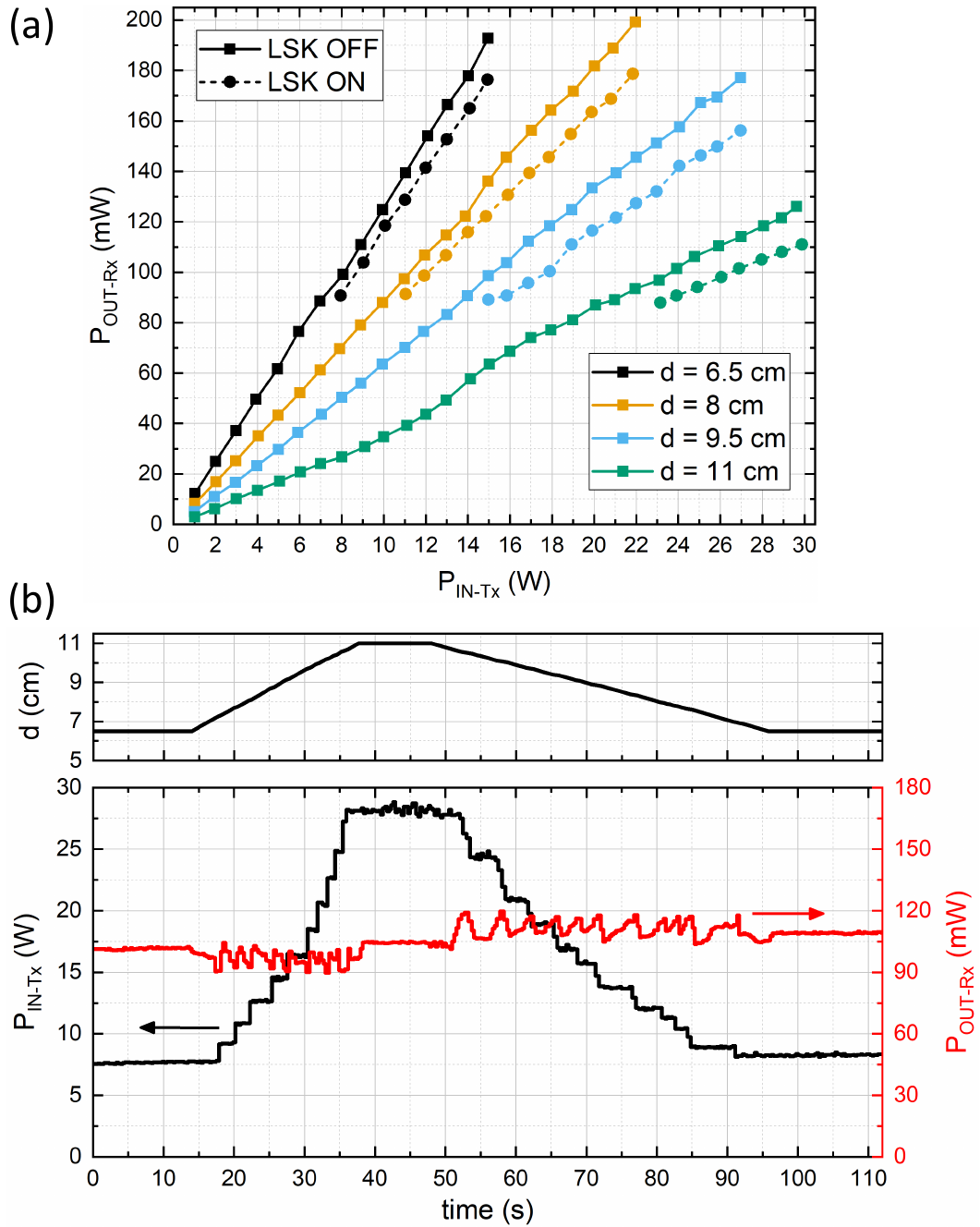}}
\caption{(a) Output power $P_{\rm{OUT-Rx}}$ with and without LSK modulation, at different input power levels $P_{\rm{IN-Tx}}$ and distances $d$ between Tx and Rx coils. (b) Transient behavior of $P_{\rm{IN-Tx}}$ and $P_{\rm{OUT-Rx}}$ resulting from the proposed adaptive control system, in response of continuous variation of $d$.}
\label{fig6}
\end{figure}

To evaluate the performance of the proposed WPT system, we measured the received power $P_{\rm{Rx}}$ at various distances $d$ and for different input power $P_{\rm{IN-Tx}}$ supplied by the APS, as shown in Fig.~\ref{fig6}(a). For the sake of simplicity, we measured the power received by the coil orthogonal to the Z-axis aligned with the Tx coil, as in the set-up shown in Fig.~\ref{fig5}(b). While the subsequent sections will illustrate the results of the received power from the 3D configuration, particularly emphasizing the effects of misalignment and rotations, here the aim is to highlight the influence of distance and the effect of LSK modulation on power efficiency. For the same targeted received power, e.g., 100~\unit{\milli\watt}, the input power must increase about three times while moving the Rx coil from a distance of 6.5~\unit{\cm} to 11~\unit{\cm} from the Tx coil. The adaptive control system addresses this problem by preventing the transmission of maximum power when the distance between the coils does not necessitate it, albeit resulting in a power efficiency reduction of around 10\% due to the impedance mismatch introduced by $C_{\rm{M}}$ during half of the modulation period.

Fig.~\ref{fig6}(b) shows a time transient test where the relative distance between the coils is dynamically changed from 6.5~\unit{\centi\meter} to 11~\unit{\centi\meter} back and forth, with a distancing and approaching speed of 0.2~\unit{\centi\meter}/\unit{\s} and 0.1~\unit{\centi\meter}/\unit{\s}, respectively. These settings are consistent with the time required for thorough investigation of the colon in capsule endoscopy procedures \cite{Xu2021}. Owing to the adaptive control system, the real-time regulation of the input power guarantees a received power close to 100~\unit{\milli\watt}. When the received power level is too low, the absence of modulation is detected by the DEM board, which increases the input voltage $V_{\rm{IN}}$ provided by the APS and consequently the received power. Likewise, when the rectifier output voltage exceeds the higher threshold, the modulation frequency shift from 15~\unit{\kilo\hertz} to 35~\unit{\kilo\hertz} is detected, leading to a reduction of $V_{\rm{IN}}$, and, consequently, of $P_{\rm{IN-Tx}}$ and $P_{\rm{OUT-Rx}}$.

\begin{figure*}[!t]
\centerline{\includegraphics[width=1\textwidth]{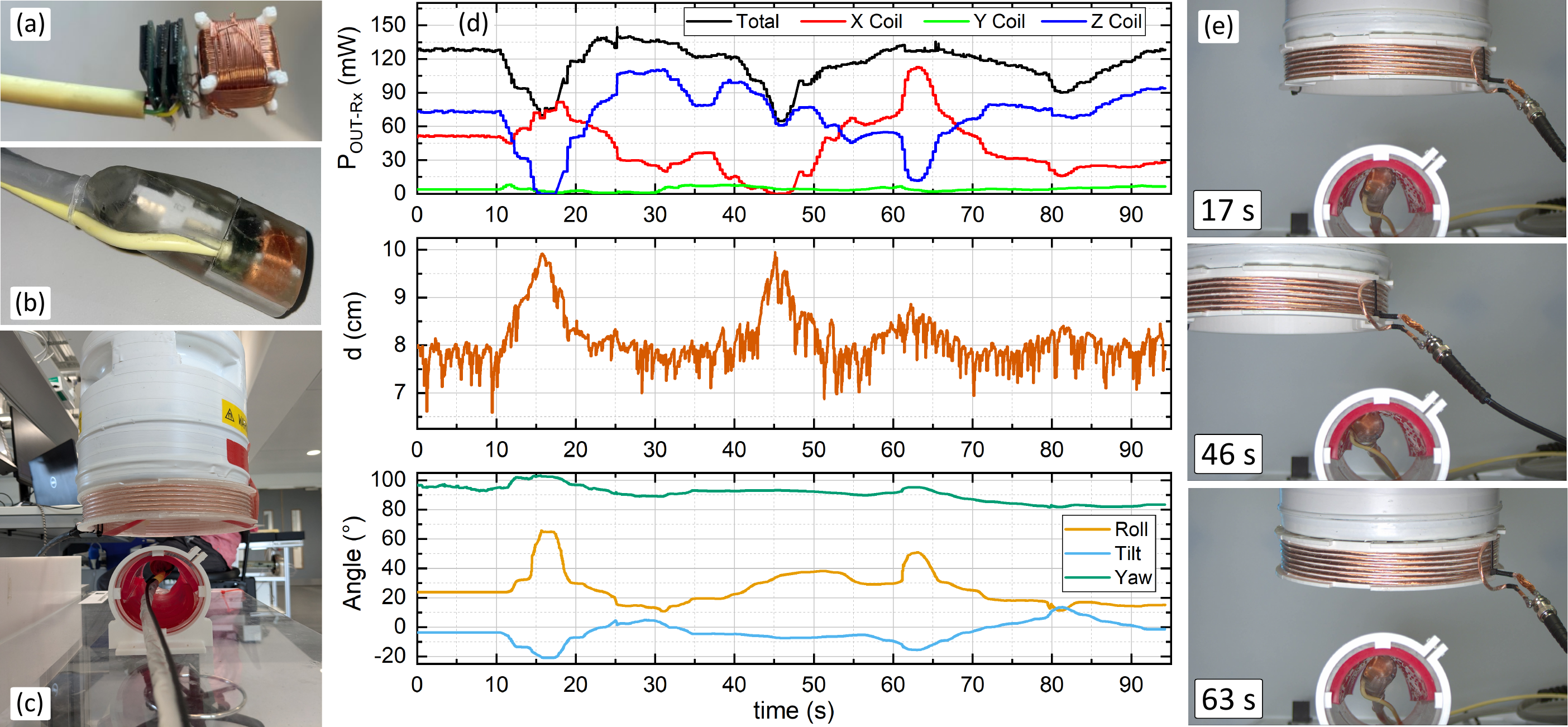}}
\caption{(a) Assembling of the WPT system and (b) its integration in the capsule demonstrator together with the IPM. (c) \textit{In vitro} setup and (d) test results in (e) different positions of the robotic arm.}
\label{fig7}
\end{figure*}

\subsection{Integration with the MCS and characterization of the robotic capsule}

During the preparation of the in vivo test, a significant effort has been devoted to the integration of the WPT receiver together with the IPM and the magnetometer inside the capsule prototype. The three miniaturized PCBs, having a shape that conforms to a capsule compartment, were stacked and closely connected to the 3D coils, as shown in Fig.~\ref{fig7}(a). The WPT receiver was then resin-coated inside the compartment (see Fig.~\ref{fig7}(b)) and put together with the IPM and the inertial and magnetic sensors. The shell of the capsule prototype was 3D-printed in resin using a Form 3B+ printer from FormLabs. The output voltages of the three rectifiers, as well as the output of the Hall-effect sensors and the IMU, are brought to the outside by means of wire connections for debugging purposes. In fact, a wired prototype without the WDT system was preferred. For the same reason, we preferred to monitor separately the three rectified output voltages ($V_{\rm{OUT-X}}$, $V_{\rm{OUT-Y}}$, and $V_{\rm{OUT-Z}}$). However, a single, more stable, supply voltage can be readily achieved by combining them as demonstrated in \cite{Pacini2018, Lee2022}. Fig.~\ref{fig7}(c) shows the setup built for the validation of the proposed robotic capsule system: the Tx coil is attached to the end effector of the KUKA LBR Med R820, a robotic manipulator with 7 degrees of freedom, while the capsule is inserted in a cylindrical tube and moved by means of the EPM inside the robotic arm. During the experiment, the output power of each rectifier (with 120-\unit{\ohm} load resistance), together with the distance between the coils and the three angles of rotation of the capsule, were monitored. Fig.~\ref{fig7}(d) shows the results of the test, under a constant input power condition ($P_{\rm{IN-Tx}} = 57~\unit{\watt}$ and with the adaptive control off).

Depending on the movements of the robotic arm, the Z-coil -- initially aligned with the Tx coil as in Fig.~\ref{fig7}(c) -- may experience suboptimal coupling with the magnetic field. This results in reduced efficiency and significant drops in output power. For instance, at 17~\unit{\s} and 63~\unit{\s}, when the robotic arm assumed the positions depicted in Fig.~\ref{fig7}(e), the power on the Z-coil approaches zero. In these situations, the magnetic coupling with the X-coil strengthens and the total power is  maintained at an adequate level due to the 3D approach. After 46~\unit{\s}, instead, the distance between the two coils increases, leading to a reduction in the power gathered by the X-coil, while the orientation of the capsule remains nearly unchanged. The adaptive control system can assist in achieving the necessary power budget in such occurences. It is also important to note that in the specific geometrical arrangement employed in this test, the Y-coil is never exploited due to its positional relationship to the Tx coil. However, in a real-case scenario, it is impossible to predict the relative orientation of the capsule and the whole 3D receiving system is needed.

\subsection{In vivo tests}

\begin{figure}[!t]
\centerline{\includegraphics[width=1\columnwidth]{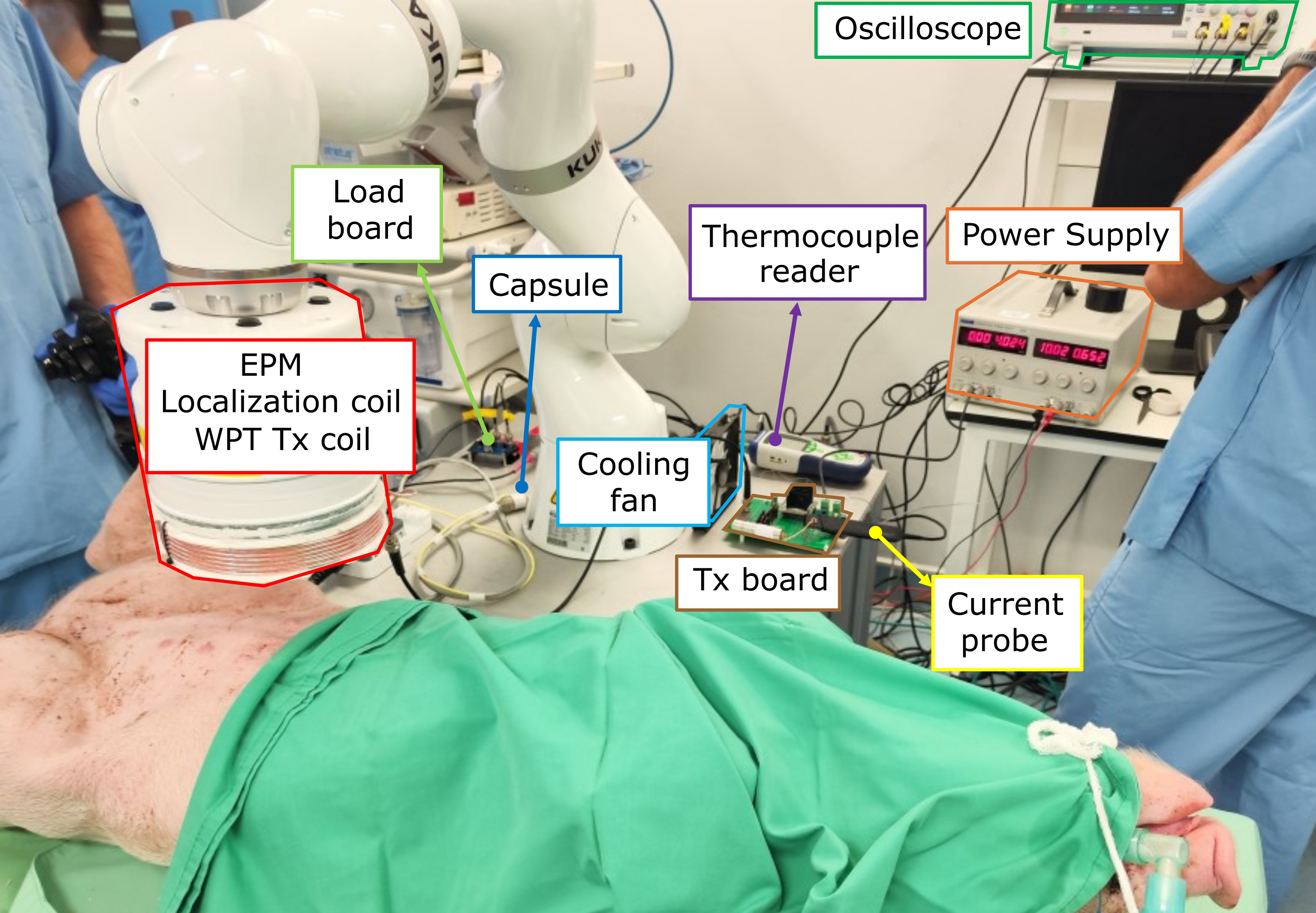}}
\caption{Picture of the setup used during \textit{in vivo} animal trials.}
\label{fig8}
\end{figure}

The animal experiments were conducted in accordance with the Animal (Scientific Procedures) Act 1986, as well as the guidelines provided by NC3Rs and ARRIVE. For a practical \textit{in vivo} demonstration of the system with clinical relevance, we selected a porcine model due to the similarity to human gastrointestinal anatomy. The primary goal of the animal trials was to validate the effectiveness and the robustness of the wireless power transfer link in realistic conditions. For this reason, at this stage of the study, the adaptive control system was not employed. Fig.~\ref{fig8} shows the setup used during the in vivo tests, with all the subsystems described in the previous Sections clearly visible. 

The Tx board is connected by means of a coaxial cable to the Tx coil attached to the end effector of the robotic arm. A thermocouple reader was placed in proximity to the power driver for real-time temperature monitoring. Thanks to the implementation of the heat sink combined with the cooling fan, the temperature remained below 60\unit{\degreeCelsius} throughout the whole measurement session. It is worth mentioning that the GaN FETs employed in the EPC90135 Half-Bridge Driver withstand an operating temperature up to 120\unit{\degreeCelsius}. The current probe Tektronix TCP0030A was employed to monitor the current flowing in the Tx coil and was connected to the oscilloscope Tektronix TBS2000B, as shown in the upper part of Fig.~\ref{fig8}. The EX354RD bench power supply was used to power the whole system: one channel provided the fixed supply voltage for the digital circuitry of the Tx board, while the second channel was employed to power the half-bridge. The picture also shows the capsule, whose rectified output voltages were connected to the remaining three oscilloscope channels. The oscilloscope was remotely controlled by means of a PC (not shown in the picture) to collect the data of the rectified voltages and the current on the Tx coil. The same PC was used also for collecting the data of the localization system and for the control of the robotic arm. 

\begin{figure}[!hb]
\centerline{\includegraphics[width=0.9\columnwidth]{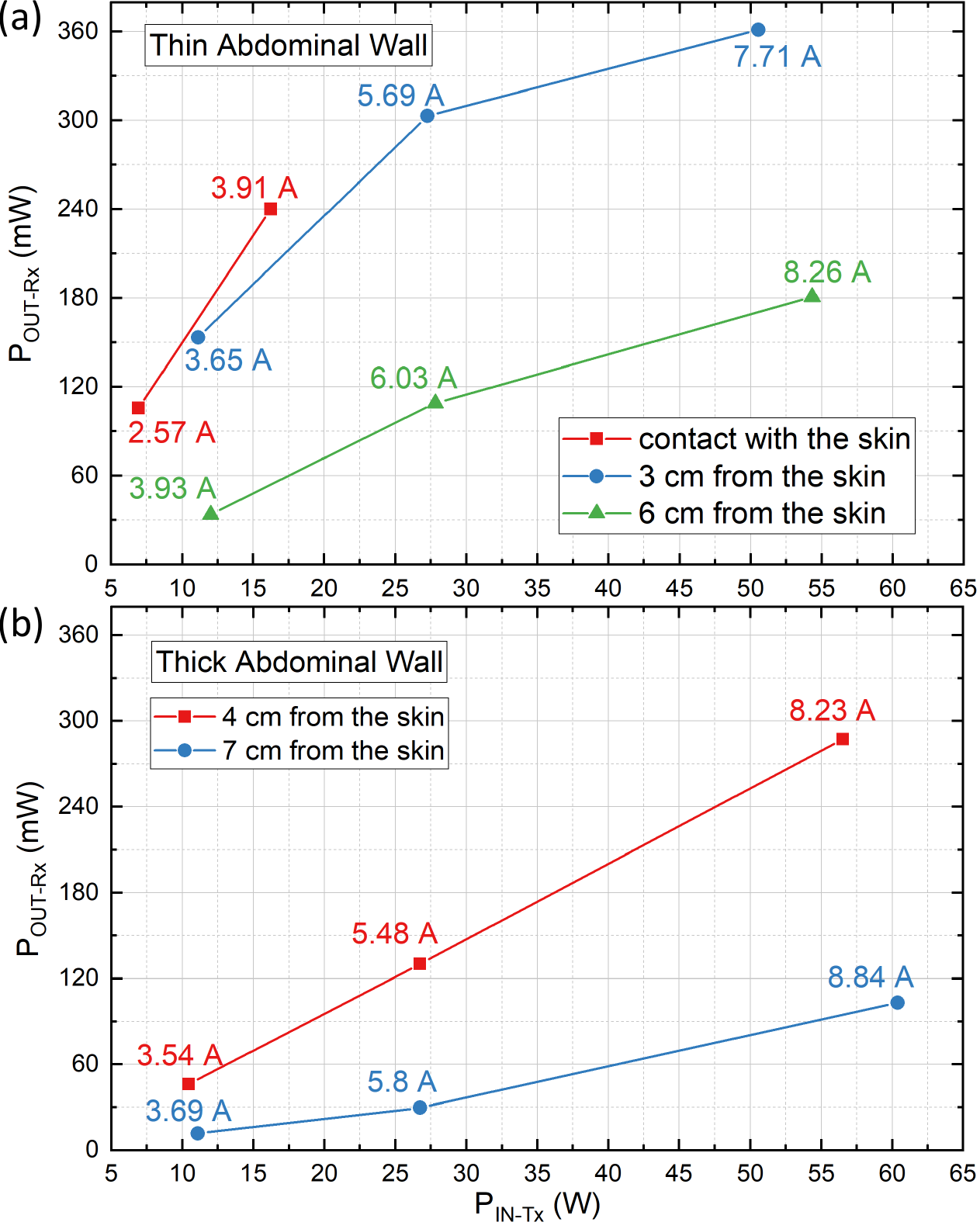}}
\caption{Results of \textit{in vivo} tests, repeated at different distances and input power levels, with static configurations of the robotic arm: (a) above thin abdominal walls, and (b) above thick abdominal wall.
% and (c) with a lateral misalignment of 10~\unit{\centi\meter} (the numbers in brackets close to the last data point of the blue and the green curves represent the actual distance only for those tests). For each point, the corresponding current on the Tx coil is shown.
}
\label{fig9}
\end{figure}

Two kinds of test have been performed during the in vivo trials: static and dynamic. Static tests were repeated in different positions (with the capsule placed below thin abdominal wall and thick abdominal wall) and also in misalignment conditions between the robotic arm and the capsule. During these tests, the capsule was kept steady by means of an auxiliary endoscope controlled manually by a physician. Each test was conducted for several levels of $P_{\rm{IN-Tx}}$ and different heights of the robotic arm. Fig.~\ref{fig9}(a) shows the average received power collected during the first tests performed in the thin abdominal wall region: in nearly all instances, except for the lowest input power and the largest coil distance, the power budget of 100~\unit{\milli\watt} was consistently achieved, with the current amplitude in the Tx coil remaining below the threshold identified for SAR constraints, as determined by electromagnetic simulations. Similar tests were repeated in the thick abdominal wall region, and the results are depicted in  Fig.~\ref{fig9}(b), showing, as expected, a significant reduction in the received power. Nonetheless, even at the largest distance, the target power budget was reached with an input power close to 60~\unit{\watt} and a current with an amplitude of 8.84~\unit{\ampere} on the Tx coil. 

%The same static tests were repeated after introducing a misalignment between the robotic arm and the capsule, as reported in Fig.~\ref{fig9}(c). In this case, the target power budget of 100~\unit{\milli\watt} was reached with low $P_{\rm{IN-Tx}}$ power when the Tx coil was in contact with the skin, medium $P_{\rm{IN-Tx}}$ at a distance of 3 \unit{\centi\meter}, and with the largest $P_{\rm{IN-Tx}}$ (54~\unit{\watt}) at 4~\unit{\centi\meter}. Even under these disadvantageous conditions, the current amplitude measured on the Tx coil remains well below the limit imposed by SAR constraints. Therefore, it is reasonable to assume that even at greater distances, an increase of the input power would have safely allowed the achievement of the target power budget. Clearly, thermal management concerns on the Tx board could arise with a prolonged increase in input voltage. Hence, additional efforts should be devoted to optimize the heat dissipation system of the Tx board.

\begin{figure}[!ht]
\centerline{\includegraphics[width=1\columnwidth]{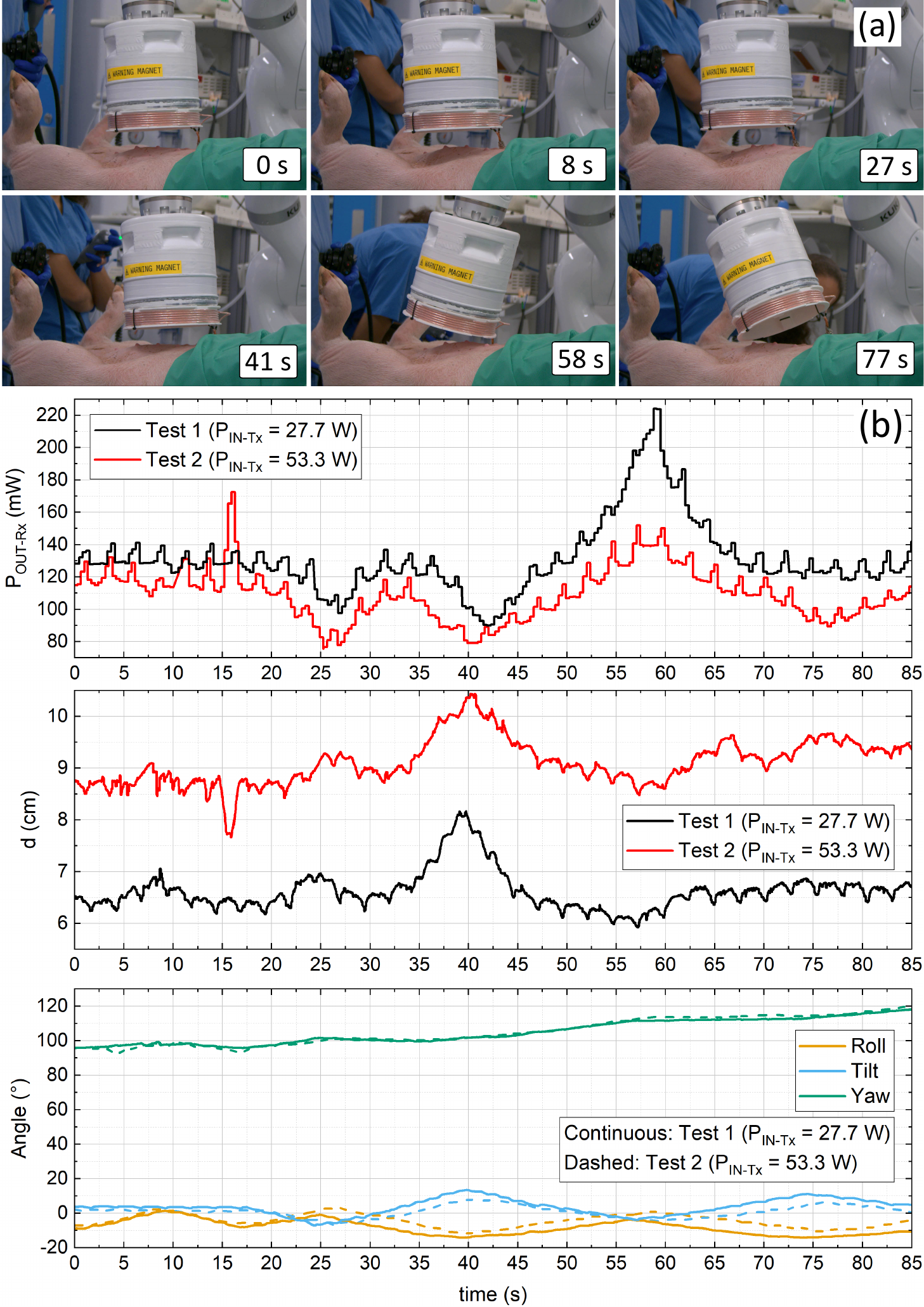}}
\caption{(a) Movements performed by the robotic arm during dynamic \textit{in vivo} tests, and (b) results of the total received power at two distances/input power levels configurations.}
\label{fig10}
\end{figure}

After the static tests, we performed dynamic tests to assess the functionality of the 3D receiver coil under different relative positions between the capsule and the robotic arm. The key snapshots of the movement pattern followed during the dynamic tests are shown in Fig.~\ref{fig10}(a), along with the corresponding timestamps. Due to the manual control of the robotic arm, slight variations in movements may occur across repeated tests. The whole test took around 85~\unit{\s} and was repeated for two combinations of robotic arm height and input power level, as shown in Fig.~\ref{fig10}(b). During Test 1, the received power was above the target power of 100~\unit{\milli\watt} except for the time intervals around 27~\unit{\s}, 41~\unit{\s} and 77~\unit{\s}, characterized by a larger distance and misalignment between the two coils. Nonetheless, even under the worst coupling conditions, $P_{\rm{OUT-Rx}}$ consistently remains over 80~\unit{\milli\watt}. At approximately 58~\unit{\s}, the distance between the coils diminishes, resulting in a peak in the received power. The implementation of the adaptive control system can restrict the variability of $P_{\rm{OUT-Rx}}$ by regulating the input power in real-time, as previously illustrated in Fig.~\ref{fig6}(b). A similar test was conducted at an increased distance (2~\unit{\centi\meter} larger) and higher input power (almost doubled) while maintaining the same movement pattern. This test confirmed the previous results and validated the robustness of the proposed WPT system in a real-case scenario. The periodic fluctuations observed in both the output power waveforms are caused by the animal's respiration. The current amplitudes on the Tx coil measured during Tests 1 and 2 remained almost constant and equal to 5.8~\unit{\ampere} and 8.2~\unit{\ampere}, respectively. These values are consistent with the results of the static tests and confirm the compliance with the SAR constraints. 

\section{Conclusion}
\label{sec:conclusion}
This work integrates wireless powering with magnetic control for the first time, marking a significant advancement toward fully autonomous, battery-free robotic capsule endoscopy systems, hence improving diagnostic capabilities in gastrointestinal healthcare.
%This paper presented the design, characterization, and in vivo validation of an inductive wireless power transfer (WPT) system for magnetically controlled robotic capsule endoscopy. 
The transmitting coil was mounted on the end effector of the robotic arm, which includes also the external permanent magnet and the localization coil for the precise capsule manipulation. This approach restricts the distance fluctuation between the transmitter and receiver, while allowing the robotic arm to move freely, unlike conventional WPT systems. By integrating a resonant inductive coupling mechanism with a 3D receiving coil inside the capsule, we demonstrated a robust power delivery solution capable of overcoming the challenges posed by coil misalignment and rotational variations. The system was further enhanced by an adaptive power control loop, leveraging load-shift keying modulation to dynamically regulate transmitted power.
Extensive laboratory testing and in vivo trials on a porcine model validated the system capability to maintain continuous and efficient power transfer under realistic clinical conditions. The results confirmed that the proposed WPT system reliably delivers sufficient power for endoscopic capsule operation while ensuring compliance with specific absorption rate safety limits and allowing precise robotic navigation via magnetic actuation. %The demonstrated approach eliminates the need for onboard batteries, addressing critical constraints in capsule endoscopy, including size, operational lifetime, and patient safety.

%Future developments will focus on further miniaturization of the WPT receiver and optimization of the inductive link.
% A single ASIC integrating the three active rectifiers required for the 3D-coil receiver will save volume while achieving higher power efficiences, while the use of a reconfigurable metasurface (e.g. designed on a conformable substrate placed over the patient's abdomen) allows a real-time spatial filtering of the magnetic field based on the position of the capsule, thus minimizing the specific absorption rate.

%\section*{Appendix}

%Appendixes, if needed, appear before the acknowledgment.

%\input{Text/bibliography}

\bibliographystyle{Bibliography/IEEEtranTIE}
%\bibliography{Text/bibliography}
\bibliography{Bibliography/IEEEabrv,Text/main}\ %IEEEabrv instead of IEEEfull

\end{document}